\documentclass{article}
\usepackage{spconf,amsmath,graphicx}
\usepackage{amsfonts}
\usepackage{float}
\usepackage{color,soul}
\usepackage{url}            
\usepackage{booktabs}

\def\vx{{\mathbf x}}
\def\vy{{\mathbf y}}

\def\va{{\mathbf a}}
\def\vb{{\mathbf b}}
\def\vi{{\mathbf i}}

\def\mX{{\mathbf X}}
\def\mA{{\mathbf A}}
\def\mB{{\mathbf B}}
\def\mC{{\mathbf C}}
\def\mD{{\mathbf D}}
\def\mW{{\mathbf W}}
\def\mQ{{\mathbf Q}}
\def\mK{{\mathbf K}}
\def\mV{{\mathbf V}}
\def\mO{{\mathbf O}}

\title{COMPLEX TRANSFORMER: A FRAMEWORK FOR MODELING COMPLEX-VALUED SEQUENCE}
\name{Muqiao Yang*, Martin Q. Ma*, Dongyu Li*, Yao-Hung Hubert Tsai, Ruslan Salakhutdinov\thanks{* equal contribution. This work was supported in part by DARPA grant FA875018C0150, DARPA SAGAMORE HR00111990016,  AFRL CogDeCON, and Apple. We would also like to acknowledge NVIDIA's GPU support.
}}
\address{Carnegie Mellon University, Pittsburgh, PA, USA \\ \texttt{\{muqiaoy, qianlim, dongyul, yaohungt, rsalakhu\}@cs.cmu.edu}}

\begin{document}
\topmargin=0mm
%
\maketitle
\begin{abstract}
While deep learning has received a surge of interest in a variety of fields in recent years, major deep learning models barely use complex numbers. However, speech, signal and audio data are naturally complex-valued after Fourier Transform, and studies have shown a potentially richer representation of complex nets. In this paper, we propose a Complex Transformer, which incorporates the transformer model as a backbone for sequence modeling; we also develop attention and encoder-decoder network operating for complex input. The model achieves state-of-the-art performance on the MusicNet dataset and an In-phase Quadrature (IQ) signal dataset. The GitHub implementation to reproduce the experimental results is available at \url{https://github.com/muqiaoy/dl_signal}.
\end{abstract}
\section{Introduction}
\label{sec:intro}
Speech recognition, signal processing, and audio transcription have been advanced by recent deep learning models \cite{amodei2016deep, oord2016wavenet}. Those models only use the real half of the spectral input. However, developing deep learning models for complex-valued input is crucial because signal based time series modeling is naturally complex-valued through Fourier Transform (FT). In recent years, complex-valued feed-forward neural network \cite{hirose2003complex, amin2009single} has been introduced, and more advanced complex-valued deep neural nets for sequence modeling have been proposed \cite{arjovsky2016unitary, danihelka2016associative, wolter2018complex, wolterfrnn}. Those methods for sequences are based on recurrent models like recurrent neural network (RNN), Long Short-Term Memory (LSTM) \cite{hochreiter1997long}, and Gated Recurrent Units (GRU) \cite{cho2014learning}, which inherently suffer from the memory bottleneck.

Recently, attention mechanism \cite{bahdanau2014neural, cho2014learning} and transformer models \cite{vaswani2017attention, yang2019xlnet} have become a well-performed sequence model because of their capability of looking at a more extended range of input contexts and representing those in different subspaces. Attention is, therefore, promising for building a complex-valued model capable of analyzing high-dimensional data with long temporal dependency. We introduce a \textit{Complex Transformer} to solve complex-valued sequence modeling tasks, including prediction and generation. The Complex Transformer adapts complex representations and operations into the transformer network. We test the model with MusicNet, a dataset for music transcription \cite{thickstun2016learning}, and a complex high-dimensional time series In-phase Quadrature (IQ) signal dataset. The specific contributions of our work are as follows: 

Our Complex Transformer achieves state-of-the-art results on the MusicNet dataset \cite{thickstun2016learning}, a dataset of classical music pieces with annotating labels at each time interval, and IQ signal dataset, a non-public dataset containing WI-FI signals with fixed-length and their corresponding device labels. Our transformer model contains several complex-valued features which boost its performance, including complex attention and complex encoder-decoder.

\section{Relation to Prior Work}
Complex-valued neural networks have attracted attentions from the deep learning community since the complex version of backpropagation algorithm \cite{benvenuto1992complex} was proposed because of its richer representational capacity. It was then applied in a variety of domains, including signal processing \cite{you1998nonlinear} and computer vision \cite{oyallon2015deep} where signals and images in their waveform or Fourier Transform are used as input data.

The work in \cite{arjovsky2016unitary, danihelka2016associative, wolter2018complex} applied complex operations on RNN, LSTM \cite{hochreiter1997long}, and GRU\cite{cho2014learning}, respectively. The work in \cite{trabelsi2017deep} implemented complex convolutional layer and used the layers to predict based on input within each time step. \cite{wolterfrnn} combined FT and sequence modeling methods to explore the temporal information. Our method differs from the previous ones as follows: Compared to \cite{arjovsky2016unitary, danihelka2016associative, wolter2018complex, wolterfrnn}, which were based on recurrent networks, our work used attention-based transformer network, which has global view of input over all time steps and do not have memory bottleneck issue embedded in backbone models of RNN, LSTM and GRU; Compared to \cite{trabelsi2017deep}, we utilize the temporal dependencies across time steps. Another important difference we made that distinguishes this work from others is that we have performed sequence generation on both acoustic input and signal input. 

\section{Model Architecture}
\label{sec:model}

Our complex transformer consists of an encoder-decoder structure, where the encoder maps a complex-valued input sequence into a complex-valued representation, and the decoder generates a complex-valued sequence one time step at a time given the encoder output.

\subsection{Problem Statement}
Given $n$ sequences of signal $(\vx_1, ..., \vx_n)$ with $d$ time steps, we use our model to classify given inputs and generate new sequences. To start with, we transform the raw real-valued signal $(\vx_1, ..., \vx_n) \in \mathbb{R}^{d \times n}$ into $(\va_1 + \vi \vb_1, ..., \va_n + \vi \vb_n) = \mA + \vi \mB \in \mathbb{C}^{d \times n}$ by Discrete Fourier Transform, an operation which decomposes a finite time sequence into a finite frequency sequence. The frequency sequence $\mX = \mA + \vi \mB$ is then fed into the transformer model. 

For an arbitrary classification task, to classify input $\mX$ to a label $\vy$, we first use a stack of encoders to produce the representation of encoder input $\mX_{\text{enc}} = \mA' + \vi \mB'$ for $\mX$, where $\mA'$ and $\mB'$ are separate latent vectors for $\mA$ and $\mB$ of $\mX$ respectively. We then use a linear layer to predict output label probability given $\mX_{\text{enc}}$. For the generation task, a stack of decoders in the model will be given both $\mX_{\text{enc}}$ and  $\mX_{\text{dec}} = \mC + \vi \mD$ (decoder input), and completes the rest of $\mX_{\text{dec}}$ by generating $\mC'$ and $\mD'$ (decoder output). 
\begin{figure*}[]
\begin{center}
\includegraphics[scale=0.332]{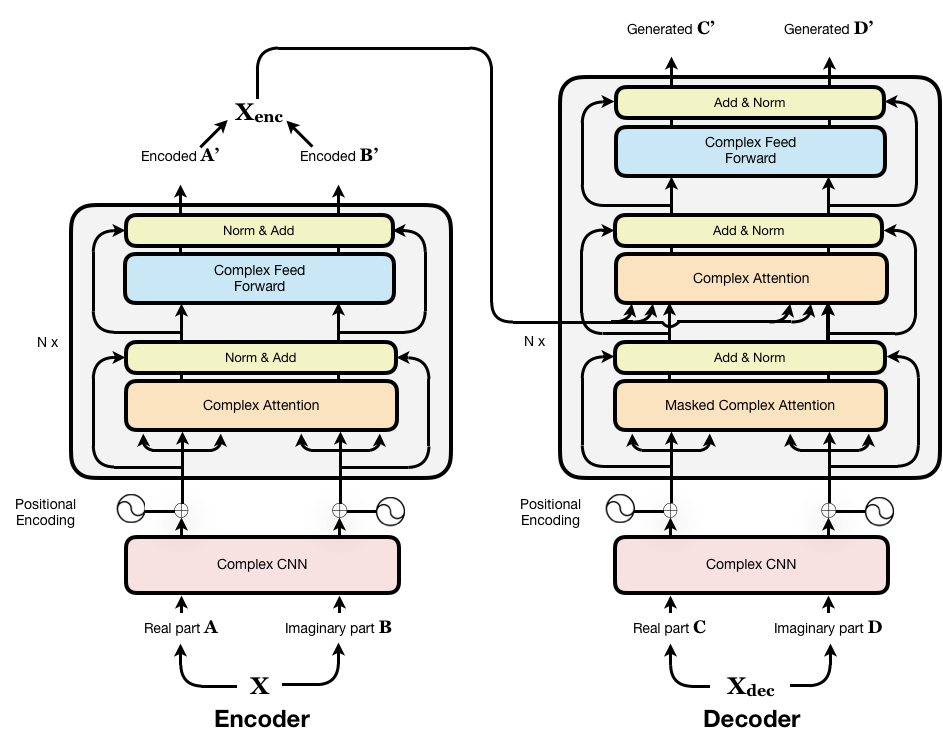}
\end{center}
\caption{Model Architecture Overview (Left: Encoder; Right: Decoder).}
\label{fig:arch}
\end{figure*}
\vspace{-3mm}

\subsection{Complex Encoder and Decoder}

Our model architecture is shown in Fig. \ref{fig:arch}. 

\textbf{Complex Encoder} The encoder is composed of six identical stacks, each with two sublayers. The first sublayer is a complex attention layer, and the second is a complex-valued feed-forward network. Both sublayers have residual connections \cite{he2016deep} and layer normalizations \cite{ba2016layer}. We employ layer normalization before residual connections in the encoder, referred as ``Norm \& Add" in the encoder part of Fig. \ref{fig:arch}.

\textbf{Complex Decoder} The decoder also has six identical stacks. Each stack has three sublayers: complex attention, complex feed-forward network, and another complex attention layer. The first complex attention layer is masked with the additional diagonal masking to prevent attending to subsequent positions. The second complex attention will be performed on the encoded representation $\mX_{\text{enc}}$ and the decoder input $\mX_{\text{dec}}$. 

\subsection{Complex Attention}

For a complex vector $\vx = \va + \vi\vb$, we represent $\va$ and $\vb$ as different input parts and simulate complex operations using real values. This is because for any complex function $f : \mathbb{C}^n \rightarrow \mathbb{C}^n $ and any complex vector $\vx = \va + \vi\vb$, we can represent $f$ as $f(\va + \vi\vb) = \alpha(\va, \vb) + \vi\beta(\va, \vb)$ where $\alpha, \beta : \mathbb{R}^n \rightarrow \mathbb{R}^n $  \cite{arjovsky2016unitary}. This indicates any complex function could be rewritten into two separate real functions.

A complex feed-forward network feeds $\va$ and $\vb$ to separate real-valued feed-forward neural networks and ReLU activations \cite{trabelsi2017deep}. A complex convolutional neural network \cite{trabelsi2017deep} convolves a complex weight matrix $\mW = \mA + \vi \mB$ and complex vector $\vx$ as the following:
\begin{equation}
\label{eq:equation}
\begin{split}
\ \ \ \ \ \mW * \vx & = (\mA + \vi \mB) * (\va + \vi \vb) \\
& = (\mA * \va - \mB * \vb) + \vi (\mA * \vb + \mB * \va)
\end{split}
\end{equation}

\begin{figure}[H]
\includegraphics[scale=0.264]{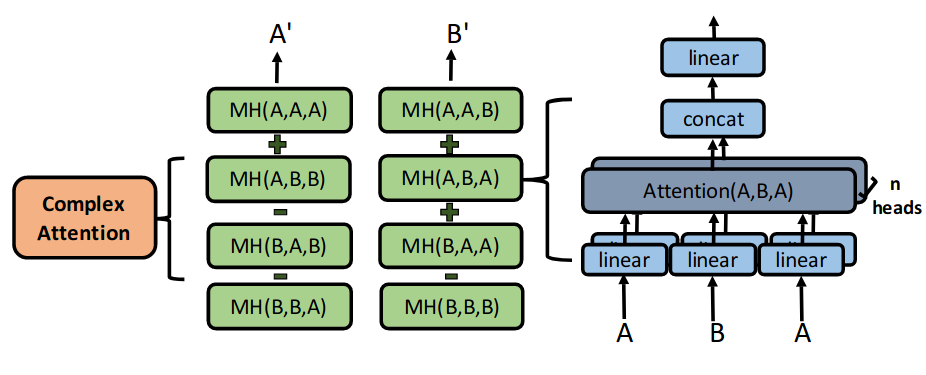}
\caption{Structure of Complex Attention.}
\label{fig:spec}
\end{figure}

Inspired by \cite{vaswani2017attention}, we propose complex building blocks for attention mechanism in our model. Given complex input $\mX = \mA + \vi \mB$, we want to attend high-dimensional information at different time steps just as in real. Thus, we compute the query matrix $\boldsymbol{\mathcal{Q}} = \mX\mW_Q$, the key matrix $\boldsymbol{\mathcal{K}} = \mX\mW_K$, and the value matrix $\boldsymbol{\mathcal{V}} = \mX\mW_V$ (where $\boldsymbol{\mathcal{Q}}, \boldsymbol{\mathcal{K}}, \boldsymbol{\mathcal{V}}$ are complex-valued and $\mW$s are real-valued) and define the complex attention: 
\begin{equation}
\label{eq:complex-attention}
\begin{split}
&\ \ \ \ \ \boldsymbol{\mathcal{Q}}\boldsymbol{\mathcal{K}}^T\boldsymbol{\mathcal{V}}  \\
& = (\mX\mW_Q)(\mX\mW_K)^T(\mX\mW_V) \\
& = (\mA\mW_Q + \vi\mB\mW_Q)(\mW_K^T\mA^T + \vi\mW_K^T\mB^T)(\mA\mW_V + \vi\mB\mW_V) \\
& = (\mA\mW_Q\mW_K^T\mA^T\mA\mW_V - \mA\mW_Q\mW_K^T\mB^T\mB\mW_V \\
&\ \ - \mB\mW_Q\mW_K^T\mA^T\mB\mW_V - \mB\mW_Q\mW_K^T\mB^T\mA\mW_V) \\ 
&\ \ + \vi(\mA\mW_Q\mW_K^T\mA^T\mB\mW_V + \mA\mW_Q\mW_K^T\mB^T\mA\mW_V \\
&\ \ + \mB\mW_Q\mW_K^T\mA^T\mA\mW_V - \mB\mW_Q\mW_K^T\mB^T\mB\mW_V) \\ 
& = \mA' + \vi \mB'
\end{split}
\end{equation}

where $\mA'$ and $\mB'$ represent the real and the imaginary part of the complex attention result respectively. In our implementation, to have a better resolution of internal similarities between the real and the imaginary parts, for each term in the expanded version of Eq. (\ref{eq:complex-attention}), we calculate the multihead attentions, as shown in Fig. \ref{fig:spec} and Eq. (\ref{eq:equation3})-Eq. (\ref{eq:equation6}).

\begin{equation}
\label{eq:equation3}
\begin{split}
&\ \ \ \ \ \text{ComplexAttention}(\mX) \\
& = (\text{MH}(\mA, \mA, \mA) - \text{MH}(\mA, \mB, \mB) \\
&\ \ - \text{MH}(\mB, \mA, \mB) -\text{MH}(\mB, \mB, \mA)) \\ 
&\ \ + \vi(\text{MH}(\mA, \mA, \mB) + \text{MH}(\mA, \mB, \mA) \\
&\ \ +\text{MH}(\mB, \mA, \mA) - \text{MH}(\mB, \mB, \mB))
\end{split}
\end{equation}
where
\begin{equation}
\begin{split}
&\ \ \ \ \ \text{MH}(\mQ, \mK, \mV) = \text{MultiHead}(\mQ, \mK, \mV) \\
& = \text{Concat}(\{\text{Attention}(\mQ\mW_i^{\mQ}, \mK\mW_i^{\mK}, \mV\mW_i^{\mV})\}_{i=1}^n)\mW^{\mO}
\end{split}
\end{equation}
\begin{equation}
    \text{Attention}(\mQ, \mK, \mV) = \text{Min-Max-Norm}(\frac{\mQ\mK^T}{\sqrt{d_k}})\mV
\end{equation}
\begin{equation}
\label{eq:equation6}
    \text{Min-Max-Norm}(\mX) = \frac{\mX - \min(\mX)}{\max(\mX) - \min(\mX)}
\end{equation}

The scaling factor $d_k$ is the feature dimension of $\mQ$ and $\mK$ (note that $\boldsymbol{\mathcal{Q/K/V}}$ are complex-valued matrices, while $\mQ/\mK/\mV$ are real-valued placeholder matrices that we could plug in $\mA$ or $\mB$). $\text{Attention}(\mQ, \mK, \mV)$ could be intuitively regarded as an extended, weight-adjusted representation of $\mV$, based on $\mV$'s dependencies on $\mQ$ and $\mK$.  $\text{MultiHead}(\mQ, \mK, \mV)$ provides a better resolution for the dependencies in different subspaces \cite{tsaikernel} (in this case we have $n$ heads, i.e. the number of attention blocks that are concatenated together). We use \textit{Min-Max-Norm} instead of \textit{Softmax} as in \cite{vaswani2017attention}, because min-max-normalization prevents gradient explosion better in our model.

\section{Experiments}

\subsection{Automatic Music Transcription}

\begin{table}
\centering
\caption{Experimental results for automatic music transcription.}
\begin{tabular}{c c c}
\toprule
 Model & \# Parameters & APS (\%)  \\
\midrule
cgRNN \cite{wolter2018complex} & 2.36M & 53.0 \\
Deep Real Network \cite{trabelsi2017deep} & 10.00M & 69.8 \\
Deep Complex Network \cite{trabelsi2017deep} & 17.14M & 72.9 \\
Concatenated Transformer & 9.79M & 71.30 \\
\textbf{Complex Transformer} & 11.61M & \textbf{74.22} \\
\bottomrule
\end{tabular}
\label{tab:result_music}
\end{table}

Automatic Music Transcription (AMT) is a challenging and significant problem, as it is a problem of forming a mapping from an audio sequence to a symbolic representation. We choose MusicNet \cite{thickstun2016learning}, a large collection of music recordings, with raw audios as data and music notes as labels at each time step. There are in total 128 labels and each time step could contain multiple labels (i.e. a multi-label classification task). We split the dataset and resampled the raw input according to \cite{trabelsi2017deep}, and then preprocess the processed signals with FT and use the complex frequency domain data as the input of our model. 

Suggested by \cite{thickstun2016learning}, we use recordings with ids ’2303’, ’2382’, ’1819’ as the test set and the remaining 327 recordings as the training set. Similar to \cite{trabelsi2017deep}, we resampled the original input from 44.1kHz to 11kHz using the technique introduced by \cite{smith2002} to improve computational efficiency. 

For Complex transformer, the number of encoders is fixed as 6, and the number of heads in multi-head attention is 8. We trained our model with an initial learning rate of $10^{-3}$ and the optimizer is Adam. The initializer of the complex transformer is Xavier uniform. The total time steps of input are 64. For dropouts, we set the dropout following self-attention to be 0, the dropout following ReLU in residual blocks to be 0.1 and the dropout of each residual block to be 0.1. 

For the networks we compare our results to, Complex Gated Recurrent Neural Network (cgRNN) \cite{wolter2018complex} and Deep Complex Network \cite{trabelsi2017deep} results are from the corresponding papers. Deep real network is a deep model treating real and imaginary parts as separate channels and concatenated these two as input. Concatenated Transformer is a vanilla transformer taking the same input as the deep real network. It is noteworthy that Deep Complex Network \cite{trabelsi2017deep} classifies the music data into 84 classes (piano notes only), while we classify data into 128 classes (both piano and other instruments notes), which is a more general task.  

The experimental results comparison is shown in Table \ref{tab:result_music}, which use average precision score (APS) as the metric. As the table shows, our complex transformer achieves a higher average precision score with much fewer parameters compared to Deep Complex Network\footnote{Parameter numbers of Deep Real Network and Deep Complex Network are based on empirical experiments using official Deep Complex Network GitHub code: \url{https://github.com/ChihebTrabelsi/deep_complex_networks}.}, and outperforms the Complex Gated Recurrent Neural Network (cgRNN) \cite{wolter2018complex} and a vanilla transformer. In terms of general deep neural network architecture, we achieve the state-of-the-art result on this dataset.

\subsection{In-phase Quadrature (IQ) Data Classification}

We also trained our model on a non-public, In-phase Quadrature (IQ) wireless signal dataset. The task is to classify the ID of the signaling device given a fixed-length sequence of WI-FI signal. This IQ dataset contains more than 103 million segments of wireless WI-FI signals from 53,853 devices (phones, laptops, tablets, etc). We resampled the data to guarantee the distribution of device IDs is uniform.

To train our model, we only consider the first 1,600 time steps of each sequence and classify each sequence into one of 1000 classes. Thus, for each data point, the input data is a vector $\in\mathbb{C}^{1600}$, and the label is a one-hot vector $\in\mathbb{R}^{1000}$. We trained our model with comparison to a feed-forward neural network, a gated recurrent network (GRU) \cite{cho2014learning} and a real-valued transformer. The experimental results are shown in Table \ref{tab:result_iq} and the Complex Transformer achieves better results than any other model we applied.

\begin{table}
\centering
\caption{Experimental results for In-phase Quadrature (IQ) data.}
\begin{tabular}{c c}
    \toprule
    Model & Accuracy (\%)  \\
    \midrule
    Feed-forward neural network & 47.12 \\
    Gated Recurrent Unit & 50.38 \\
    Concatenated Transformer & 57.57 \\
    \textbf{Complex Transformer} & \textbf{59.94} \\
    \bottomrule
\end{tabular}
\label{tab:result_iq}
\end{table}

\subsection{Music and IQ Dataset Conditional Generation}


Following the idea of conditional generation in \cite{huang2018music}, we generate sequences conditioned on partial input. We split the input $\mX$ into two parts. The first 60\% time steps of $\mX$ are input for the encoder. After getting a representation $\mX' = \mA' + \vi \mB'$ from the encoder input above, we then use $\mX'$ and the masked, remaining 40\% time steps of $\mX$ (masked decoder input) to generate an output sequence one step at a time. We perform this conditional generation on MusicNet and the IQ dataset and compare our model with the LSTM encoder-decoder model \cite{hochreiter1997long} as well as a vanilla transformer. Both the LSTM and the vanilla transformer concatenates the real and the imaginary parts of the signal and then take it as input. Lastly, all models use prediction layers to predict labels based on generated sequences. 

We choose binary cross entropy loss for MusicNet and cross entropy loss for IQ, since MusicNet is a multi-label classification while IQ is multi-class. As Table \ref{tab:result_gen} shows, our complex transformer has outperformed LSTM and concatenated transformer in terms of the loss between the predicted labels and ground truth.

\begin{table}[H]
    \centering
    \caption{Experimental results (in loss) for sequence generation.}
    \begin{tabular}{c c c}
        \toprule
        Model & MusicNet & IQ dataset   \\
        \midrule
        LSTM & 0.0629 & 3.7086 \\
        Concatenated Transformer & 0.0509 & 2.2580 \\
        \textbf{Complex Transformer} & \textbf{0.0492} & \textbf{2.2335} \\
        \bottomrule
    \end{tabular}
    \label{tab:result_gen}
\end{table}

\section{Conclusions}

The Complex Transformer we have introduced shows the power of sequence modeling in the complex domain. The model uses attention to capture dependencies between the real and the imaginary part in different time steps, and achieves better performance on automatic music transcription and generation, as well as signal prediction and generation tasks than other sequential models. By introducing complex operations to the attention network architecture, we show complex numbers capable of capturing richer temporal information.


\vfill\pagebreak

\bibliographystyle{IEEEbib}
\bibliography{strings,refs}

\end{document}